\documentclass[11pt]{article} 
\usepackage{url}
\usepackage{smile}
\usepackage{graphicx} 
\usepackage{todonotes}
\usepackage{epstopdf}
\usepackage{wrapfig}
\usepackage[colorlinks, linkcolor=blue, anchorcolor=blue, citecolor=blue]{hyperref}
\usepackage[margin=1in]{geometry}
\usepackage[normalem]{ulem}
\usepackage[export]{adjustbox}
\usepackage{mathtools, cuted}
\usepackage{natbib}
\usepackage{enumerate}
\usepackage{microtype}
\usepackage{algorithm}
\usepackage{booktabs} 
\usepackage{smile}
\usepackage{wrapfig}
\usepackage{xspace}
\newcommand{\ie}{\emph{i.e.}\xspace} 
\newcommand{\eg}{\emph{e.g.}\xspace} 

\usepackage{kpfonts}
\DeclareMathAlphabet{\mathsf}{OT1}{cmss}{m}{n}
\SetMathAlphabet{\mathsf}{bold}{OT1}{cmss}{bx}{n}

\begin{document}

\title{Calibrated Language Model Fine-Tuning for In- and Out-of-Distribution Data }

\author{Lingkai Kong, Haoming Jiang, Yuchen Zhuang, Jie Lyu, Tuo Zhao, Chao Zhang \thanks{All authors are affiliated with Georgia Institute of Technology.  Emails: \{\tt lkkong,jianghm, yczhuang,jie.lyu,tourzhao,chaozhang\}@gatech.edu.}}
\date{}

\maketitle

\begin{abstract}
Fine-tuned pre-trained language models can suffer from severe miscalibration for both in-distribution and out-of-distribution (OOD) data due to
over-parameterization. To mitigate this issue, we propose a regularized fine-tuning method. Our method introduces two types of regularization for
better calibration: (1) \textit{On-manifold regularization}, which generates pseudo on-manifold samples through interpolation within the data
manifold. Augmented training with these pseudo samples imposes a smoothness regularization to improve in-distribution calibration. (2)
\textit{Off-manifold regularization}, which encourages the model to output uniform distributions for pseudo off-manifold samples to address the
over-confidence issue for OOD data. Our experiments demonstrate that the proposed method outperforms existing calibration methods for text
classification in terms of expectation calibration error, misclassification detection, and OOD detection on six datasets. Our code can be found at \url{https://github.com/Lingkai-Kong/Calibrated-BERT-Fine-Tuning}.
\end{abstract}

\section{Introduction}

Pre-trained language models have recently brought the natural language processing (NLP) community into the transfer
learning era. The transfer learning framework consists of two stages, where we first pre-train a large-scale language model, (\eg, BERT \citep{devlin2019bert}, RoBERTa \citep{liu2019roberta}, ALBERT \citep{Lan2020ALBERT} and T5 \citep{raffel2019exploring})
on a large text corpus and then fine-tune it on downstream tasks.
Such a fine-tuning approach has achieved
SOTA performance in many NLP benchmarks \citep{wang2018glue, wang2019superglue}. 

\begin{figure}[h]
\centering
\includegraphics[width=0.5\linewidth]{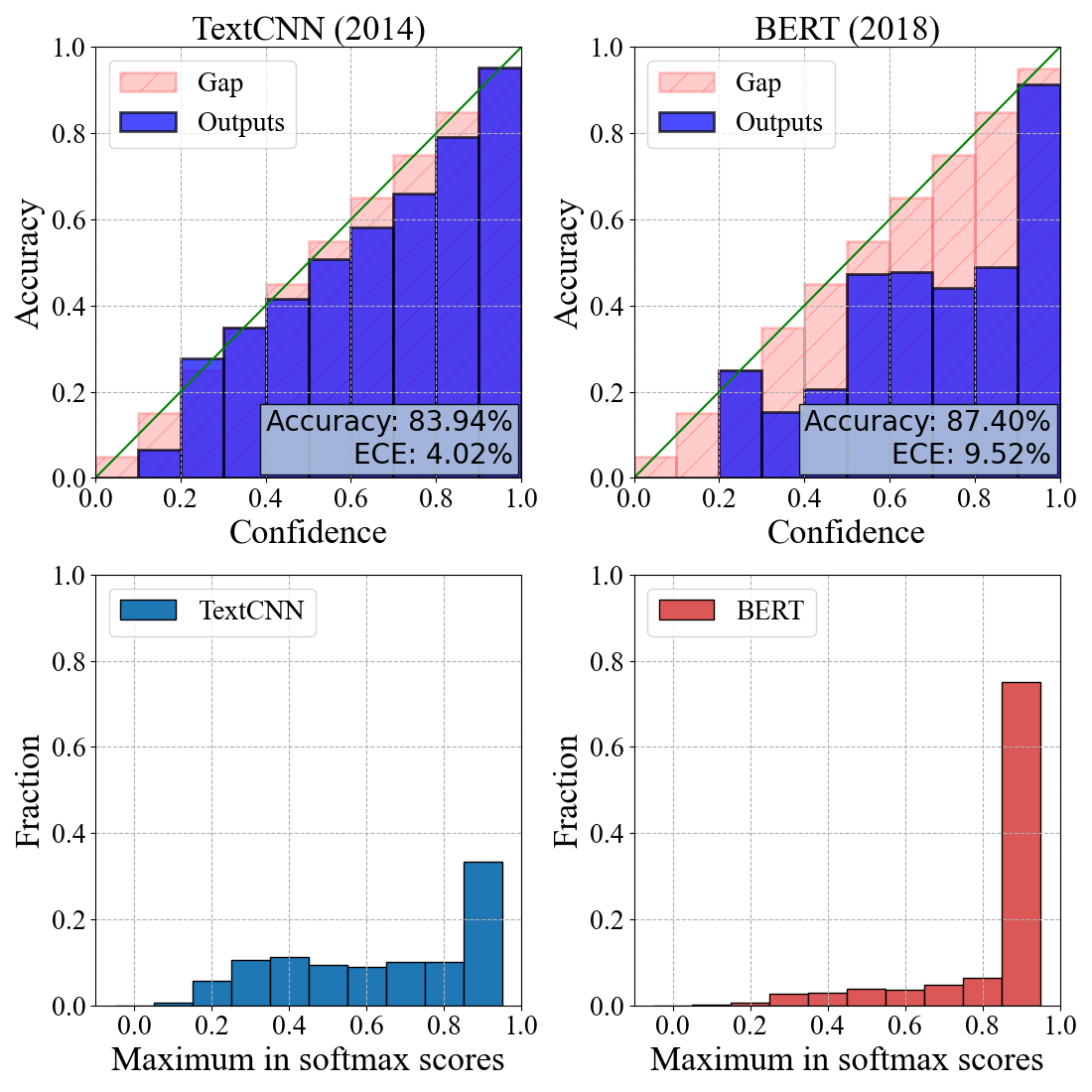}
\caption{ The reliability diagrams on in-distribution data (the first row) and the histograms of the model confidence on out-of-distribution (OOD) data (the second row) of CNN \citep{kim2014convolutional} and fine-tuned BERT-MLP classifier \citep{devlin2019bert}. Though BERT improves classification accuracy, it makes  over-confident predictions for both in-distribution and OOD data.} 
\label{fig:intro-calibration}
\end{figure}

Many applications, however, require trustworthy predictions that need to be not only accurate but also well \textit{calibrated}.
In particular, a well-calibrated model should produce reliable confident estimates for both in-distribution and out-of-distribution (OOD) data: (1) For in-distribution data, a model should produce predictive probabilities close to the true likelihood for each class, \ie, confidence $\approx$ true likelihood.
(2) For OOD data, which do not belong to any class of the training data, the model output should produce high uncertainty to say `I don't know', \ie, confidence $\approx$ random guess, instead of producing absurdly wrong yet wildly confident predictions. Providing such calibrated output probabilities can help us to achieve better model robustness \citep{lee2018simple}, model fairness \citep{chouldechova2017fair} and improve label efficiency via uncertainty driven learning \citep{gal2017active,siddhant2018deep, shen2018deep}.

Unfortunately, \citet{guo2017calibration} have shown that due to over-parameterization, deep convolutional neural networks are often  miscalibrated.
Our experimental investigation further corroborates that fine-tuned language models can suffer from miscalibration even more for NLP tasks. As shown in
  Figure~\ref{fig:intro-calibration}, we present the calibration of a BERT-MLP model for a text classification task on the 20NG dataset.
Specifically, we train a TextCNN \citep{kim2014convolutional} and a BERT-MLP using 20NG\textsubscript{15}
(the first 15 categories of 20NG) and then evaluate them on both in-distribution and
OOD data.
The first row plots their reliability diagrams \citep{mizil-2005-predict} on the test set of
20NG\textsubscript{15}. Though BERT improves the classification accuracy from $83.9\%$ to
$87.4\%$, it also increases the expected calibration error (ECE, see more details in Section \ref{sec:pre}) from $4.0\%$ to
$9.5\%$.  This indicates that BERT-MLP is much more miscalibrated for in-distribution
  data. The second row plots the histograms of the model confidence, \ie, the maximum output probability,
on the test set of 20NG\textsubscript{5} (the unseen 5 categories of 20NG).  While it is desirable to produce low probabilities for
  these unseen classes, BERT-MLP produces wrong yet over-confident
  predictions for such OOD data.

Such an aggravation of miscalibration is due to the even more significant over-parameterization of these language models.
At the pre-training stage, they are trained on a huge amount of unlabeled data in an unsupervised manner, \eg, T5 is pre-trained on 745 GB text. To
capture rich semantic and syntactic information from such a large corpus, the language models are designed to have enormous capacity, \eg, T5 has
about 11 billion parameters. At the fine-tuning stage, however, only limited labeled data are available in the downstream tasks. With the extremely
high capacity, these models can easily overfit training data likelihood and be over-confident in their predictions.

To fight against miscalibration, a natural option is to apply a calibration  method such as temperature scaling \citep{guo2017calibration} in a post-processing step. However, temperature scaling only learns a single parameter to rescale all the logits, which is not flexible and insufficient. Moreover, it cannot improve out-of-distribution calibration.
A second option is to mitigate miscalibration during training using regularization. For example, \citet{Pereyra2017erl} propose an entropy regularizer to prevent over-confidence, but it can needlessly hurt legitimate high confident predictions. A third option is to use Bayesian neural networks \citep{pmlr-v37-blundell15, pmlr-v70-louizos17a}, which treat model parameters as probability distributions to represent model uncertainty explicitly. However, these Bayesian approaches are often prohibitive, as the priors of the model parameters are difficult to specify, and exact inference is intractable, which can also lead to unreliable uncertainty estimates.

 We propose a regularization approach  to addressing miscalibration for
  fine-tuning pre-trained language models
 from a data augmentation perspective.
 We propose two new regularizers using pseudo samples both \textit{on} and
 \textit{off} the data manifold to mitigate
 data scarcity and prevent over-confident predictions. 
   Specifically, our method imposes two types of regularization for better calibration during fine-tuning:
(1) \textbf{On-manifold regularization}: We first generate  \textit{on-manifold samples} by interpolating the training data and their corresponding
labels along the direction learned from hidden feature space; training over such augmented on-manifold data introduces a smoothness constraint within
the data manifold to improve the model calibration for in-distribution data.
(2) \textbf{Off-manifold regularization}: We generate  \textit{off-manifold} samples by adding relatively large perturbations along the directions that point outward the data manifold;
we penalize the negative entropy of the output distribution for such off-manifold samples to address the over-confidence issue for OOD data.

We evaluate our proposed model calibration method on six text classification datasets. For in-distribution data, we measure ECE and the performance of misclassification detection. For out-of-distribution data, we measure the performance of OOD detection. Our experiments show that our method
outperforms existing state-of-the-art methods in both settings, and meanwhile maintains competitive classification accuracy.

We summarize our contribution as follows: (1) We propose a general calibration framework, which can be applied to pre-trained language model fine-tuning, as well as other deep neural network-based prediction problems. (2) The proposed method adopts on- and off-manifold regularization from a data augmentation perspective to improve calibration for both in-distribution and OOD data. (3) We conduct comprehensive experiments showing that our method outperforms existing calibration methods in terms of ECE, miscalssification detection and OOD detection on six text classification datasets.

\section{Preliminaries}
\label{sec:pre}

We describe model calibration for both in-distribution and out-of-distribution data.

\noindent\textbf{Calibration for In-distribution Data:}
For in-distribution data, a well-calibrated model is expected to output prediction
confidence comparable to its classification accuracy. For example, given 100
data points with their prediction confidence $0.6$, we expect $60$ of them to be
correctly classified. More precisely, for a data point $X$, we denote by $Y(X)$
the ground truth label, ${\hat Y}(X)$ the label predicted by the model, and $\hat{P}(X)$
  the output probability associated with the predicted label. The calibration error of the predictive model for a given confidence $p\in(0,1)$ is defined as:
\begin{equation}
    \mathcal{E}_p=|\mathbb{P}(\hat{Y}(X)=Y(X)|\hat{P}(X)=p) - p|.
    \label{eq:id}
\end{equation}
As \eqref{eq:id} involves population quantities,
we usually adopt empirical approximations \citep{guo2017calibration} to estimate the calibration error. Specifically, we partition all data points into $M$ bins of equal size according to their prediction confidences. Let $\mathcal{B}_m$ denote the bin with prediction confidences bounded between $\ell_m$ and $u_m$. Then, for any $p\in[\ell_m,u_m)$, we define the empirical calibration error as:
\begin{equation}
\hat{\mathcal{E}}_p=\hat{\mathcal{E}}_m=\frac{1}{|\mathcal{B}_m|}\Big|\sum_{i\in\mathcal{B}_m}\big[\mathbf{1}(\hat{y}_i=y_i)-\hat{p}_i\big]\Big|,
\end{equation}
where $y_i$, $\hat{y}_i$ and $\hat{p}_i$ are the true label, predicted label and confidence for sample $i$.

To evaluate the overall calibration error of the predictive model, we can futher take a weighted average of the calibration errors of all bins, which is also known as the expected calibration error (ECE) \citep{naeini2015obtaining} defined as:
\begin{align}
    {\rm ECE} =\sum_{m=1}^M\frac{|\mathcal{B}_m|}{n} \hat{\mathcal{E}}_{m},
    \label{eq:ece}
\end{align}
where $n$ is the sample size.

We remark that the goal of calibration is to minimize the calibration error without significantly sacrificing prediction accuracy. Otherwise, a random guess classifier can achieve zero calibration error.

\noindent\textbf{Calibration for Out-of-distribution Data:}
In real applications, a model can encounter test data that significantly differ
from the training data. For example, they come from other unseen classes, or
they are potential outliers. A well-calibrated model is expected to produce an
output with high uncertainty for such out-of-distribution (OOD) data, formally,
\begin{align*}
P(Y=j) = 1/K\quad\forall j = 1,...,K,
\end{align*}
where $K$ is the number of classes of the training data. As such, we can detect OOD data by setting up an uncertainty threshold.

\section{Calibrated Fine-Tuning via Manifold Smoothing}
\label{sec:method}
 \begin{figure*}[t]
  \centering
   \includegraphics[width=0.75\linewidth]{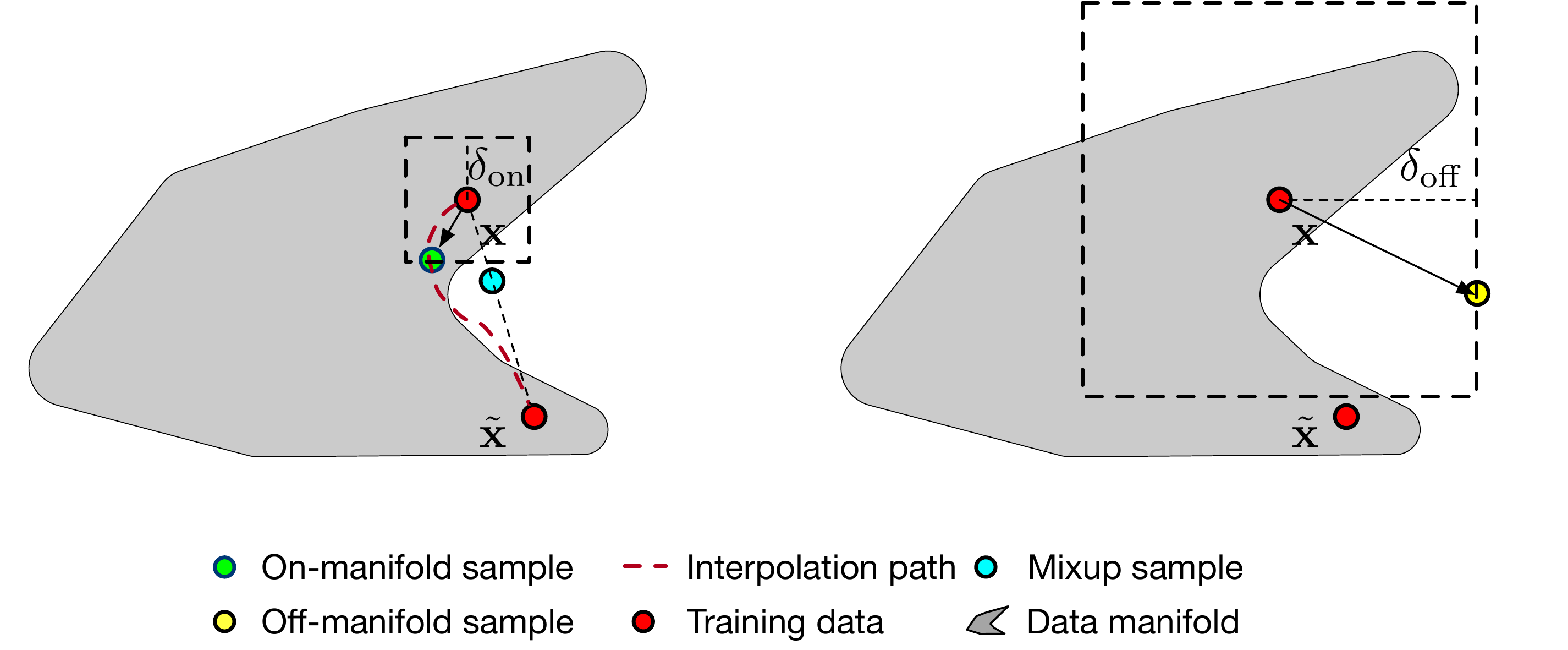}
  \caption{The on-manifold and off-manifold samples generated by our calibration procedure. Mixup adopts a coarse linear interpolation and the generated data point may deviate from the data manifold.}
 \label{fig:manifold}
\end{figure*}

 We consider $N$ data points of the target task $S
=\{(\mathbf{x}_i,y_i)\}_{i=1}^N$, where $\mathbf{x}_i$'s denote the input embedding of
the sentence and $y_i$'s are the associated
one-hot labels. Let $f(\cdot)$ denote the feature extraction layers (\eg, BERT); let $g(\cdot)$ denote the task-specific layer; and let $\theta$ denote all parameters of $f$ and $g$. We propose to optimize the following objective at the fine-tuning stage:
\begin{align}
    \min_{\theta}\mathcal{F}(\theta) = ~\mathbb{E}_{\mathbf{x},y \sim S} \ell(g \circ f(\mathbf{x}),y)
    +\lambda_{\rm on} \mathcal{R}_{\rm on}(g\circ f) + \lambda_{\rm off} \mathcal{R}_{\rm off}(g\circ f),
    \label{eq:objective}
\end{align}
where $\ell$ is the cross entropy loss, and $\lambda_{\rm on}, \lambda_{\rm off}$
are two hyper-parameters. The regularizers $\mathcal{R}_{\rm on}$ and $\mathcal{R}_{\rm off}$ are for on- and off-manifold calibration, respectively.

\subsection{On-manifold Regularization}

The on-manifold regularizer $\mathcal{R}_{\rm on}$ exploits the interpolation of training data within the data manifold to improve the in-distribution calibration. Specifically, given two training samples $(\mathbf{x}, y)$ and $(\tilde{\mathbf{x}}, \tilde{y})$ and the feature extraction layers $f$, we generate an on-manifold pseudo sample $(\mathbf{x}^{\prime},y^{\prime})$ as follows:
\begin{align}
    \mathbf{x}^{\prime *} &= \argmin_{\mathbf{x}^{\prime} \in \mathbb{B}(\mathbf{x}, \delta_{\rm on})}D_x(f(\mathbf{x}^{\prime}), f(\tilde{\mathbf{x}})), \label{eq:in-domain:x}\\
    y^{\prime} & = (1-\delta_y)y+\delta_y\tilde{y},
\label{eq:in-domain:y}
\end{align}
where $\delta_{\rm on}$ and $\delta_y$ are small interpolation parameters for data and label, and $D_x$ is a proper distance for features extracted by $f$ such as cosine distance, \ie, $D_x(\mathbf{a},\mathbf{b})=\langle \mathbf{a}/\|\mathbf{a}\|_2,\mathbf{b}/\|\mathbf{b}\|_2\rangle$, and 
$\mathbb{B}(\mathbf{x}, \delta_{\rm on})$ denotes an $\ell_{\infty}$ ball centered at $\mathbf{x}$ with a radius $\delta_{\rm on}$, \ie, $$\mathbb{B}(\mathbf{x}, \delta_{\rm on}) = \{\mathbf{x}'~|~\|\mathbf{x}'-\mathbf{x}\|_{\infty}\leq\delta_{\rm on}\}.$$

As can be seen, $\mathbf{x}^{\prime *}$ is essentially interpolating between $\mathbf{x}$ and $\tilde{\mathbf{x}}$ on the data manifold, and $D_{x}(f(\cdot),f(\cdot))$ can be viewed as a metric over such a manifold. However, as
$f(\cdot)$ is learnt from finite training data, it can recover the actual data manifold only up to a certain statistical error. Therefore, we constrain $\mathbf{x}^{\prime *}$ to stay in a small neighborhood of $\mathbf{x}$, which ensures $\mathbf{x}^{\prime *}$ to stay close to the actual data manifold.

This is different from existing interpolation methods such as Mixup \citep{zhang2018mixup, verma-2019-manifold}. These methods adopt coarse linear interpolations either in the input space or latent feature space, and the generated data may significantly deviate from the data manifold.

Note that our method not only interpolates $\mathbf{x}$ but also $y$. This can yield a soft label for $\mathbf{x}'^{*}$, when $\mathbf{x}$ and $\tilde{\mathbf{x}}$ belong to different classes. Such an interpolation is analogous to semi-supervised learning, where soft pseudo labels are generated for the unlabelled data. These soft-labelled data essentially induce a smoothing effect, and prevent the model from making overconfident predictions toward one single class.

We remark that our method is more adaptive than the label smoothing method \citep{muller2019does}. As each interpolated data point involves at most two classes, it is unnecessary to distribute probability mass to other classes in the soft label. In contrast, label smoothing is more rigid and enforces all classes to have equally nonzero probability mass in the soft label.

We then define the on-manifold regularizer as
\vspace{-0.03in}
\begin{align*}
    \mathcal{R}_{\rm on} (g\circ f) = \mathbb{E}_{(\mathbf{x}^{\prime},y^{\prime}) \sim S_{\rm on}} D_{\rm KL}(y^{\prime},g \circ f(\mathbf{x}^{\prime})),
\end{align*}
\vspace{-0.03in}
where $S_{\rm on}$ denotes the set of all pseudo labelled data generated by our interpolation method, and $D_{\rm KL}$ denotes the KL-divergence between two probability simplices.

\begin{algorithm}[t]
    \caption{Our Proposed Efficient Stochastic Optimization Algorithm for Solving \eqref{eq:objective}. $d$ is the dimension of features.}
    \begin{algorithmic}
         \For{$\#$ training iterations}
        \State Sample a mini-batch  $B=\{\mathbf{x}_i,y_i\}$ from $S$.
        \State \textit{// Generate on-manifold samples:}

        \State For each $\mathbf{x}_i \in B$, randomly select $\{\tilde{\mathbf{x}}_i,\tilde{y}_i\}$ from $B$,
        initialize $\mathbf{x}^{\prime}_i  \leftarrow \mathbf{x}_i + v_i$ with $v_i \sim \textrm{UNIF}[-\delta_{\rm on}, \delta_{\rm on}]^d$
         \State $\Delta^{\prime}_i\leftarrow {\rm sign} (\nabla_{\mathbf{x}^{\prime}_i}D_x(f(\mathbf{x}_i^{\prime}),f(\tilde{\mathbf{x}}_i)))$
            \State $\mathbf{x}_i^{\prime}\leftarrow \Pi_{\|\mathbf{x}^{\prime}_i-\mathbf{x}_i\|_{\infty}\leq \delta_{\rm on}}
            (\mathbf{x}^{\prime}_i-\delta_{\rm on} \Delta^{\prime}_i)$
            \State $y^{\prime} \leftarrow (1-\delta_y)y_i+\delta_y \tilde{y}_i$
        \State \textit{// Generate off-manifold samples:}
        \State For each $\mathbf{x}_i \in B$, initialize $\mathbf{x}^{\prime\prime}_i  \leftarrow \mathbf{x}_i + v^{\prime}_i$ with $v^{\prime}_i \sim \textrm{UNIF}[-\delta_{\rm off}, \delta_{\rm off}]^d$  \State $\Delta^{\prime\prime}_i\leftarrow {\rm sign} (\nabla_{\mathbf{x}^{\prime\prime}_i}
        \ell(g\circ f(\mathbf{x}_i^{\prime\prime}), y)$
            \State $\mathbf{x}_i^{\prime\prime}\leftarrow \Pi_{\|\mathbf{x}^{\prime\prime}_i-\mathbf{x}_i\|_{\infty}= \delta_{\rm off}}
            (\mathbf{x}^{\prime\prime}_i+\delta_{\rm off} \Delta^{\prime\prime}_i)$
        \State Update $\theta$ using ADAM
\EndFor
    \end{algorithmic}
    \label{alg:training}
\end{algorithm}

\subsection{Off-manifold Regularization}
\vspace{-0.03in}
The off-manifold regularizer, $\mathcal{R}_{2}$, encourages the model to yield low confidence outputs for samples outside the data manifold, and thus mitigates the over-confidence issue for out-of-distribution (OOD) data. Specifically, given a training sample $(\mathbf{x}, y)$, we generate an off-manifold pseudo sample $\mathbf{x}^{''\ast}$ by:
\begin{align}
    \mathbf{x}^{''\ast}=\max_{\mathbf{x}''\in \mathbb{S}(\mathbf{x},\delta_{\rm off})} \!\!\!\!\ell(g\circ f(\mathbf{x}^{\prime\prime}), y),
    \label{eq:out-of-domain}
\end{align}
where $\mathbb{S}(\mathbf{x},\delta_{\rm off})$ denotes an $\ell_{\infty}$ sphere centered at $\mathbf{x}$ with a radius $\delta_{\rm off}$.

Since we expect $\mathbf{x}^{\prime\prime\ast}$ to mimic OOD data, we first need to choose a relatively large $\delta_{\rm off}$ such that the sphere $\mathbb{S}(\mathbf{x},\delta_{\rm off})$ can reach outside the data manifold.
Then, we generate the pseudo off-manifold sample from the sphere along the adversarial direction. Existing literature \citep{stutz2019disentangling, gilmer2018relationship} has shown that such an adversarial direction points outward the data manifold.

By penalizing the prediction confidence for these off-manifold samples, 
we are able to encourage low prediction confidence for OOD data. Hence, we define the off-manifold regularizer as
\begin{align}
    \mathcal{R}_{\rm off}(g \circ f)=\mathbb{E}_{\mathbf{x}^{''} \sim S_{\rm off}} \!\!-\mathcal{H}(g\circ f(\mathbf{x}^{\prime\prime})),
\end{align}
where $S_{\rm off}$ denotes the set of all generated off-manifold samples, and $\mathcal{H}(\cdot)$ denotes the entropy of the probability simplex.

\subsection{Model Training}
We can adopt stochastic gradient-type algorithms such as ADAM \citep{kingma2014adam} to optimize \eqref{eq:objective}. At each iteration, we need to first solve two inner optimization problems in \eqref{eq:in-domain:x} and \eqref{eq:out-of-domain}, and then plug $\mathbf{x}^{\prime}$ and $\mathbf{x}^{\prime\prime}$ into \eqref{eq:objective} to compute the stochastic gradient. The two inner problems can be solved using the projected sign gradient update for multiple steps. In practice, we observe that one  single update step with random initialization is already sufficient to efficiently optimize $\theta$. Such a phenomenon has also been observed in existing literature on adversarial training \citep{wong2020fast}. We summarize the overall training procedure in Algorithm \ref{alg:training}.

\section{Experiments}
\label{sec:exp}

To evaluate calibration performance for in-distribution data, we measure the expected calibration error (ECE) and the misclassification detection score. For out-of-distribution data, we measure the OOD detection score.

We detect the misclassified and OOD samples by model confidence, which is the output probability associated with the predicted label $\hat P(X)$. Specifically, we setup a confidence threshold $\tau \in [0,1]$, and take the samples with confidence below the threshold, \ie, $\hat P(X) < \tau$, as the misclassified or OOD samples.
We can compute the detection $F_1$ score for every $\tau$: $F_1(\tau)$, and obtain a calibration curve ($F_1(\tau)$ vs. $\tau$).
Then, we set $\tau_{\rm upper}$ as the upper bound of the confidence threshold, since a well calibrated model should provide probabilities that reflect the true likelihood  and it is not reasonable to use a large $\tau$ to detect them.
We use the empirical Normalized Bounded Area Under the Calibration Curve (NBAUCC) as the overall detection score:
\begin{align*}
 {\rm NBAUCC}_{\tau_{\rm upper}}=\frac{1}{M}\sum_{i=1}^{M}  F_1\large\left(\frac{\tau_{\rm upper}}{M}i\large\right),
\end{align*}
where $M$ is the number of sub-intervals for the numerical integration. We set $M=50$ throughout the following experiments.
Note that the traditional binary classification metrics, \eg, AUROC and AUPR, cannot measure the true calibration because the model can still achieve high scores even though it has high confidences for the misclassified and OOD samples. We provide more explanations of the metrics in Appendix~\ref{sec:appendix:metric}.
We report the performance when $\tau_{\rm upper} =0.5$ here and the results when $\tau_{\rm upper} =0.7$ and $1$ in Appendix~\ref{sec:appendix:tau}.

\subsection{Datasets}

For each dataset, we construct an in-distribution training set, an in-distribution testing set, and an OOD testing set. Specifically,
we use the
following datasets:

\noindent \textbf{20NG\footnote{We use the 20 Newsgroups dataset from: \url{http://qwone.com/~jason/20Newsgroups/}}}. The 20 Newsgroups dataset (20NG) contains news articles with 20 categories. We use Stanford Sentiment Treebank (SST-2) \citep{socher2012deep} as the OOD data.

\noindent \textbf{20NG\textsubscript{15}}. We take the first 15 categories of 20NG as the in-distribution data and the other 5 categories (20NG\textsubscript{5}) as the OOD data.

\noindent \textbf{WOS} \citep{kowsari2017HDLTex}. Web of Science (WOS) dataset contains scientific articles with 134 categories. We use AGnews \citep{zhang2015character} as the OOD data.

\noindent \textbf{WOS\textsubscript{100}}. We use the first 100 classes of WOS as the in-distribution data and the other 34 classes (WOS\textsubscript{34}) as the OOD data.

\noindent \textbf{Yahoo} \citep{chang2008importance}. This dataset contains questions with 10 categories posted to `Yahoo! Answers'. We randomly draw $2000$ from $140,000$ samples for each category as the training set. We use Yelp \citep{zhang2015character} as the OOD data.

\noindent \textbf{Yahoo\textsubscript{8}}. We use the first 8 classes of Yahoo as the in-distribution data and the other 2 classes (Yahoo\textsubscript{2}) as the OOD data.

The testing set of OOD detection consists of the in-distribution testing set and the
OOD data. More dataset  details  can  be found in Appendix \ref{sec:app:dataset}. We remark that 20NG\textsubscript{15}, WOS\textsubscript{100}, and
Yahoo\textsubscript{8} are included to make OOD detection more challenging, as the OOD data and the training data come from similar data sources.

\subsection{Baselines}
We consider the following baselines:

\noindent$\bullet$ $\textbf{BERT}$ \citep{devlin2019bert} is a pre-trained base BERT model stacked with one linear layer.\\
\noindent$\bullet$ \textbf{Temperature Scaling (TS)} \citep{guo2017calibration} is a
post-processing calibration method that learns a single parameter to rescale the logits on the development set after the model is fine-tuned. \\
\noindent$\bullet$ \textbf{Monte Carlo Dropout (MCDP)} \citep{gal2016dropout} applies dropout at testing time for multiple times and then averages the outputs. \\
\noindent$\bullet$ \textbf{Label Smoothing (LS)} \citep{muller2019does} smoothes the one-hot label by distributing a certain probability mass to other non ground-truth classes. \\
\noindent$\bullet$ \textbf{Entropy Regularized Loss (ERL)} \citep{Pereyra2017erl} adds a entropy penalty term to prevent DNNs from being over-confident. \\
\noindent$\bullet$ \textbf{Virtual Adversarial Training (VAT)} \citep{miyato2018virtual} introduces a smoothness-inducing adversarial regularizer to encourage the local Lipschitz continuity of DNNs.\\
\noindent$\bullet$ \textbf{Mixup} \citep{zhang2018mixup,sunil2019mixup} augments training data by linearly interpolating training samples in the input space. \\
\noindent$\bullet$ \textbf{Manifold-mixup (M-mixup)} \citep{verma-2019-manifold} is an extension of Mixup, which interpolates training samples in the hidden feature space.

\subsection{Implementation Details}
We use ADAM \citep{kingma2014adam} with $\beta_1=0.9$ and $\beta_2=0.999$ as the optimizer.
For our method, we simply set $\lambda_{\rm on}=\lambda_{\rm off}=1, \delta_{\rm on} =10^{-4}, \delta_{\rm off}=10^{-3}$, and $\delta_y=0.1$ for all the experiments.
We also conduct an extensive hyper-parameter search for the baselines.  See more details in Appendix~\ref{app:exp_details}.

\subsection{Main Results}
\label{sec:exp-main}

\begin{figure}[t]
\centering
\includegraphics[width=0.7\linewidth]{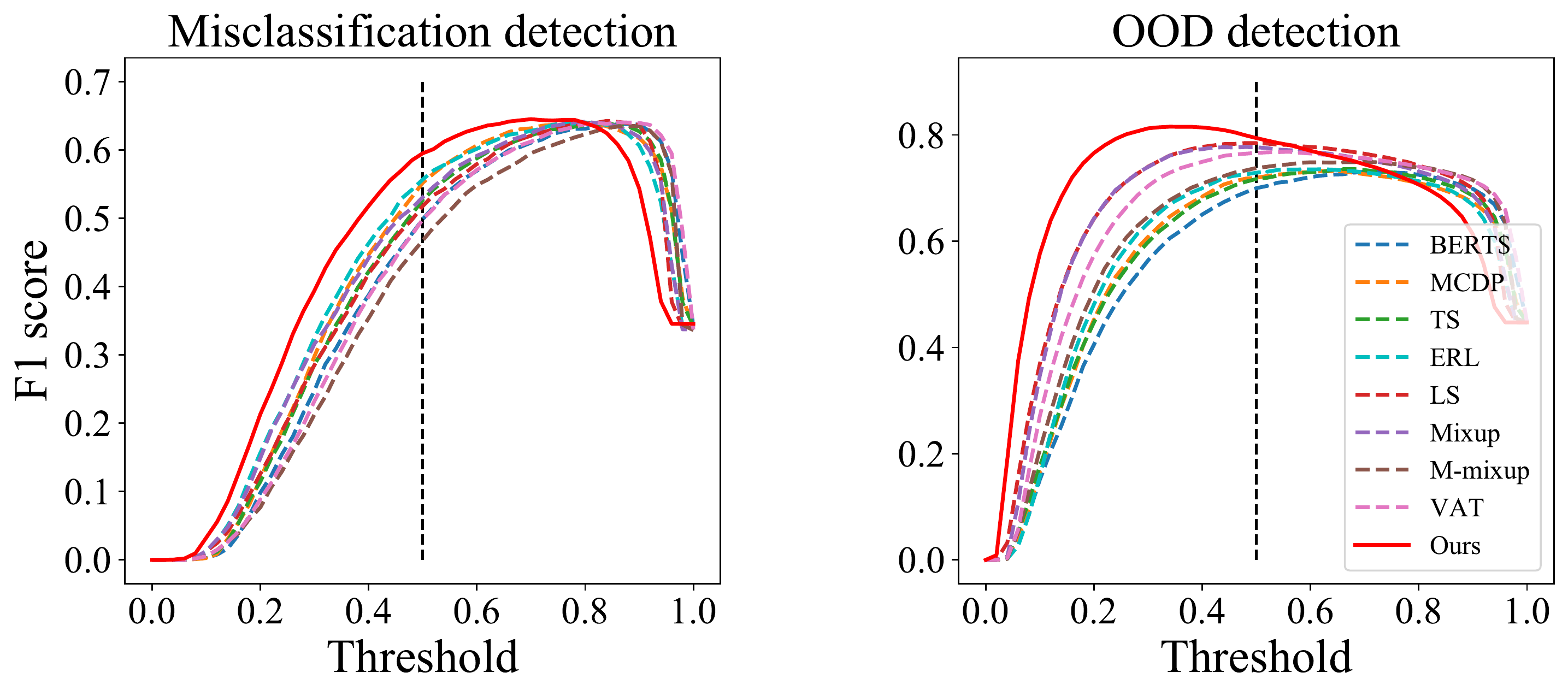}
\caption{ Calibration curves of OOD detection and misclassification detection on WOS. Our method can achieve high $F_1$ scores starting from a small threshold which indicates that it indeed provides low confidences for misclassified and OOD samples; the $F_1$ scores of the baselines peak at high thresholds which indicates that they are poorly calibrated. } 
\vspace{-0.1in}
\label{fig:cc}
\end{figure}

Our main results are summarized as follows:

\noindent\textbf{Expected Calibration Error}: Table~\ref{table:ece} reports the ECE and predictive accuracy of all the methods. Our method outperforms all the baselines on all the datasets in terms of ECE except for Yahoo, where only ERL is slightly better. Meanwhile, our method does not sacrifice the predictive accuracy.

\noindent\textbf{Misclassification Detection:}
Table~\ref{table:mis_ood} compares the ${\rm NBAUCC}_{0.5}$ on misclassification detection of different methods. As shown, our method
outperforms all the baselines on all the six datasets.

\noindent\textbf{Out-of-distribution Detection:} Table~\ref{table:mis_ood} reports the ${\rm NBAUCC}_{0.5}$ on OOD detection of different methods. Again, our method achieves the best
performance on all the six datasets. The improvement is particularly remarkable
on the 20NG dataset, where ${\rm NBAUCC}_{0.5}$ increases from $47.00$ to $63.92$ compared with
the strongest baseline. We also find that detecting the unseen classes from the
original dataset is much more challenging than detecting OOD samples from a totally different dataset.

\noindent\textbf{Significance Test:}
We perform the Wilcoxon signed rank test \citep{wilcoxon1992individual} for significance test. For each dataset, we conduct experiments using 5 different random seeds with significance level $\alpha=0.5$. We find that our model outperforms other baselines on all the datasets significantly, with only exceptions of ERL in ECE on Yahoo and ERL in misclassification detection on 20NG.

\begin{table*}[t]
\centering
\begin{tabular}{@{}l@{}c@{~}c@{~}c@{~}c@{~}c@{~}c@{~}|@{~}c@{~}c@{~}c@{~}c@{~}c@{~}c@{}}
\toprule
\multirow{2}{*}{Model} & \multicolumn{6}{@{}c@{~}|@{~}}{ECE} & \multicolumn{6}{c}{Accuracy} \\ \cline{2-13}
 & 20NG\textsubscript{15} & 20NG & WOS\textsubscript{100} & WOS & Yahoo\textsubscript{8} & Yahoo & 20NG\textsubscript{15} & 20NG & WOS\textsubscript{100} & WOS & Yahoo\textsubscript{8} & Yahoo \\ \hline
BERT & $9.24$ & $11.61$ & $6.81$ & $6.74$ & $10.11$ & $10.54$ & $87.42$ & $84.55$ & $81.94$ & $79.40$ & $73.58$ & $71.89$ \\
TS & $4.42$ & $8.17$ & $3.63$ & $4.43$ & $5.18$ & $4.24$ & $87.42$ & $84.55$ & $81.94$ & $79.40$ & $73.58$ & $71.89$ \\
MCDP & $6.88$ & $9.17$ & $4.00$ & $3.55$ & $6.54$ & $6.72$ & $87.45$ & $84.55$ & {$82.09$} & {$79.67$} & $73.67$ & $71.99$ \\
LS & $4.35$ & $6.15$ & $4.35$ & $4.67$ & $4.89$ & {$3.61$} & $87.54$ & {$85.02$} & $81.95$ & $79.47$ & $73.66$ & $71.54$ \\
ERL & $7.16$ & $6.10$ & $3.74$ & $3.35$ & $3.42$ &{$\bm{2.96}$} & $87.67$ & $84.83$ & $81.96$ & $79.48$ & $73.63$ & $72.01 $\\
VAT & $9.07$ & $11.28$ & $7.27$ & $6.76$ & $10.96$ & $7.92$ & $87.61$ & $85.20$ & $81.65$ & $79.71$ & $73.71$ & $72.08$ \\

Mixup & $5.98$ & $9.02$ & $4.72$ & $4.21$ & $4.60$ & $5.18$ & $87.49$ & $84.86$ & $81.97$ & $79.51$ & $73.88$ & $71.82$ \\
M-mixup & $5.04$ & $7.78$ & $6.48$ & $6.68$ & $7.01$ & $6.07$ & $87.40$ & $84.45$ & $81.77$ & $79.57$ & $73.67$ & $72.03$ \\ \hline
\textbf{Ours} & \textbf{3.69} & \textbf{4.43} & \textbf{3.24} & \textbf{3.04} & \textbf{3.03} & 3.42 & $87.44$ & $84.53$ & $81.59$ & $79.06$ & {$73.71$} & ${72.17}$ \\ \bottomrule
\end{tabular}
\caption{ECE and accuracy (in percentage). We report the average performance of 5 random initializations.}
\label{table:ece}
\end{table*}

\begin{table*}[!h]
\centering
\begin{tabular}{@{}l@{}c@{~}c@{~}c@{~}c@{~}c@{~}c@{~}|@{~}c@{~}@{~}c@{~}c@{~}c@{~}c@{~}c@{~}c@{~}c@{}}
\toprule
&& \multicolumn{4}{c}{Misclassification Detection} && \multicolumn{6}{c}{OOD Detection}
\\
\cline{2-13}

Data & \multirow{2}{*}{20NG\textsubscript{15}} & \multirow{2}{*}{20NG} & \multirow{2}{*}{WOS\textsubscript{100}} & \multirow{2}{*}{WOS} & \multirow{2}{*}{Yahoo\textsubscript{8}} & \multirow{2}{*}{Yahoo} 
& 20NG\textsubscript{15} & 20NG & WOS\textsubscript{100} & WOS & Yahoo\textsubscript{8} & Yahoo \\
 
( OOD )&  &  &  &  &  &  
& 20NG\textsubscript{5} & SST-2 & WOS\textsubscript{34} & AGnews & Yahoo\textsubscript{2} & Yelp \\ \hline
BERT
& $2.30$ & $2.86$  & $16.53$ & $20.52$ & $7.47$ & $8.43$
& $2.66$ & $21.65$ & $23.12$ & $49.84$ & $8.35$ & $13.88$\\ 
TS 
& $6.08$ & $5.74$ & $21.20$ & $23.76$ & $10.48$ & $12.74$ 
& $6.62$ & $32.64$ & $28.12$ & $53.32$ & $11.55$ & $20.27$\\ 
MCDP
& $4.37$ & $5.28$ & $20.44$ & $24.16$ & $10.12$ & $10.75$ 
& $3.99$ & $25.10$ & $27.28$ & $53.52$ & $9.98$ & $15.93$\\ 
LS 
& $4.72$ & $6.75$ & $20.37$ & $23.56$ & $11.19$ & $16.15$ 
& $5.70$ & $41.08$ & $27.12$ & $58.48$ & $12.02$ & $19.81$\\ 
ERL
& $8.54$ & $10.35$ & $20.49$ & $25.13$ & $12.89$ & $15.47$ 
& $8.78$ & $47.00$ & $27.73$ & $56.67$ & $13.78$ & $23.47$\\ 
VAT 
& $2.52$ & $3.36$ & $18.70$ & $19.96$ & $6.54$ & $10.37$ 
& $2.96$ & $29.62$ & $23.41$ & $54.60$ & $7.42$ & $17.65$\\ 
Mixup 
& $4.99$ & $4.51$ & $20.65$ & $24.80$ & $10.75$ & $11.29$ 
& $5.86$ & $31.84$ & $26.77$ & $58.02$ & $11.62$ & $19.84$\\ 
M-mixup 
& $2.16$ & $3.16$ & $16.94$ & $19.39$ & $9.09$ & $11.79$ 
& $2.36$ & $26.08$ & $24.08$ & $51.39$ & $10.08$ & $22.41$\\ 
\hline
\textbf{Ours} 
& \textbf{9.10} & \textbf{10.76} & \textbf{26.93} & \textbf{30.80} & \textbf{14.34} & \textbf{17.88}
& \textbf{9.69} & \textbf{63.92} & \textbf{35.60} & \textbf{71.13} & \textbf{14.94} & \textbf{29.40} 
\\ \bottomrule
\end{tabular}
\caption{${\rm NBAUCC}_{0.5}$ on misclassification detection and OOD detection. We report the average performance of 5 random initializations.}
\label{table:mis_ood}
\end{table*}

\subsection{Parameter Study}
\label{sec:parameter-study}
\begin{figure*}[t]
 \centering
 \includegraphics[width=\linewidth]{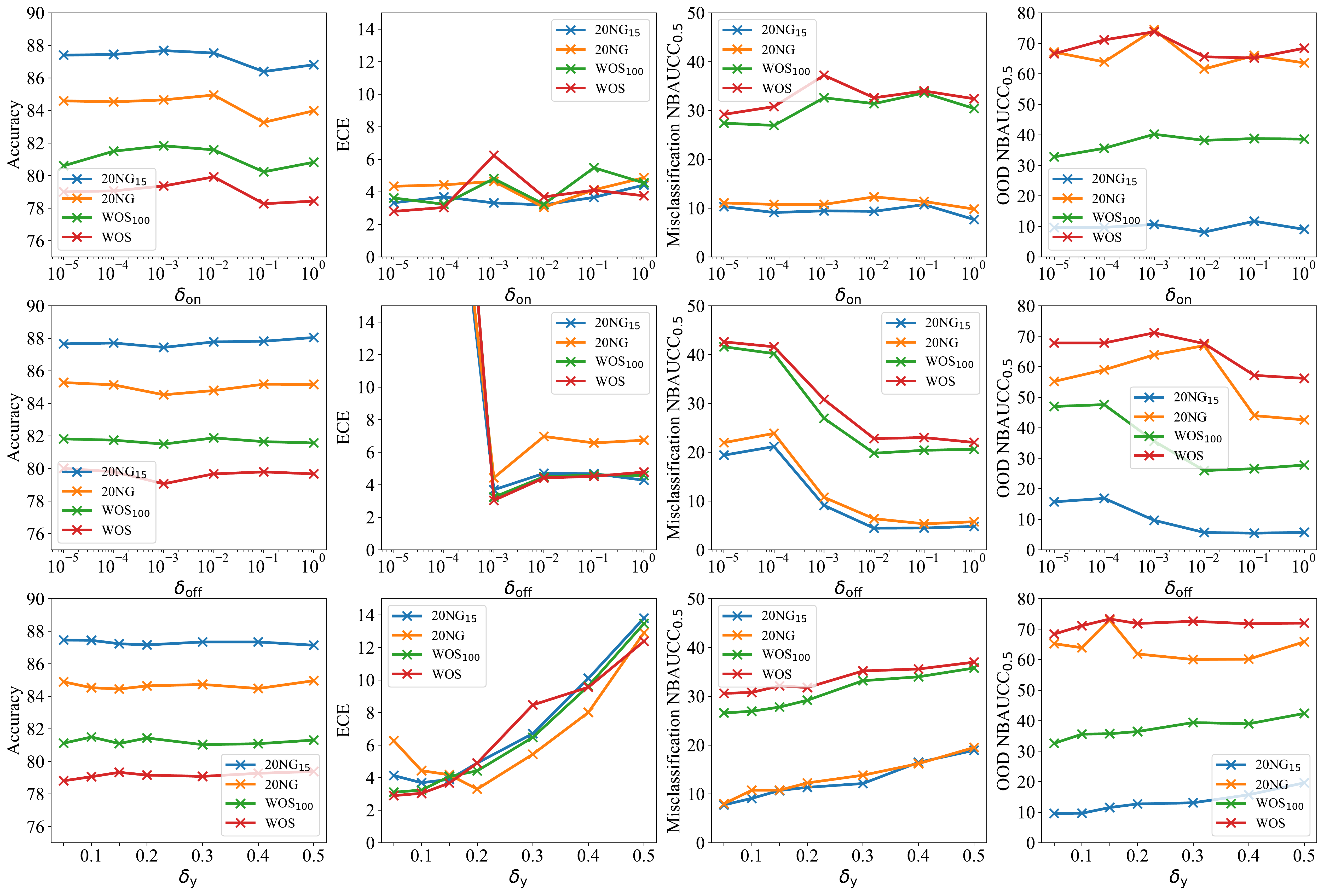}
 \caption{ Parameter study of $\delta_{\rm on}$, $\delta_{\rm off}$ and $\delta_y$. }
 \label{fig:para1}
\end{figure*}


We investigate the effects of the interpolation parameters for on-manifold data, \ie, $\delta_{\rm on}$ and $\delta_y$, and the perturbation size for off-manifold samples, \ie, $\delta_{\rm off}$.
The default values are $\delta_{\rm on}=10^{-4}, \delta_{\rm off} = 10^{-3}$ and $\delta_y=0.1$.
Figure~\ref{fig:para1} shows the reuslts on 20NG\textsubscript{15}, 20NG, WOS\textsubscript{100}, and WOS datasets. Our results are summarized as follows:

\noindent$\bullet$~The performance of all metrics versus $\delta_{\rm on}$ is stable within a large range from $10^{-5}$ to $10^{-2}$. When $\delta_{\rm on}$ is larger than $10^{-1}$, the predictive accuracy begins to drop.

\noindent$\bullet$~The performance versus $\delta_{\rm off}$ is more sensitive: (1) when $\delta_{\rm off}$ is too small, ECE increases dramatically becasue the generated off-manifold samples are too close to the manifold and make the model under-confident. (2) when $\delta_{\rm off}$ is too large, the off-manifold regularization is too weak and OOD detection performance drops.

\noindent$\bullet$~In general, $\delta_{\rm on}$ should be small to let $\mathbf{x}^{\prime}$ stay on the data manifold while $\delta_{\rm off}$ should be large to let $\mathbf{x}^{\prime\prime}$ leave the data manifold.
However, the regularization effect of $\mathcal{R}_{\rm on}$ ($\mathcal{R}_{\rm off}$) depends on both $\lambda_{\rm on}$ ($\lambda_{\rm off}$) and $\delta_{\rm on}$ ($\delta_{\rm off}$). Therefore, it is not necessary to let $\delta_{\rm on}$ be smaller than $\delta_{\rm off}$.
We can also tune $\lambda_{\rm on}$ and $\lambda_{\rm off}$ to achieve better performance.

\noindent$\bullet$~The performance versus $\delta_y$ is relatively stable except for the metric of ECE. When $\delta_y$ is larger than $0.2$, ECE begins to increase.

\subsection{Ablation Study}
We investigate the effectiveness of the on-manifold regularizer
$\mathcal{R}_{\rm on}$ and the off-manifold regularizer $\mathcal{R}_{\rm off}$
via ablation studies. Table~\ref{table:ablation} shows the results on the  20NG\textsubscript{15} and 20NG datasets.

\noindent$\bullet$~As expected, removing either component in our method would
result in a performance drop. It demonstrates that these two components
complement each other. All the ablation models outperform the BERT baseline
model, which demonstrates the effectiveness of each module.

\noindent$\bullet$~We observe that the optimal $\delta_{\rm on}$ is different when using only $\mathcal{R}_{\rm on}$. This indicates that the hyperparameters of $\mathcal{R}_{\rm on}$ and $\mathcal{R}_{\rm off}$ should be jointly tuned, due to the joint effect of both components.

\noindent$\bullet$~By removing $\mathcal{R}_{\rm off}$, we observe a severe OOD performance degradation on the 20NG dataset (from 63.92 to 43.87). This indicates
that $\mathcal{R}_{\rm off}$ is vital to out-of-distribution calibration. Meanwhile, the performance degradation is less severe on 20NG\textsubscript{15} (from 9.69 to 7.94). It is because $\mathcal{R}_{\rm on}$ can also help detect the OOD samples from similar data sources. (20NG\textsubscript{5}).

\noindent$\bullet$~By removing $\mathcal{R}_{\rm on}$, the in-distribution calibration performance drops as expected.

\begin{table*}[!htb]
\centering
\begin{tabular}{lc|cccc|cccc}
\toprule
 \multicolumn{2}{c|}{Dataset}& \multicolumn{4}{c|}{20NG\textsubscript{15}} & \multicolumn{4}{c}{20NG} \\ \hline
Model &$\delta_{\rm on}$ & Accuracy & ECE & OOD & Mis & Accuracy & ECE & OOD & Mis \\ \hline
BERT & -   & $87.42$ & $9.24$ & $2.66$ & $2.30$
& $84.55$ & $11.61$ & $21.65$ & $2.86$\\ \hline
w/ $\mathcal{R}_{\rm off}$ & - & $86.48$ & $6.51$ & $6.22$ & $6.09$
& $83.90$ & $7.98$ & $55.40$ & $7.12$ \\ \hline
w/ $\mathcal{R}_{\rm on}$  & $10^{-2}$   & $88.73$ & $2.77$ & $7.94$  & $8.08$  
& $85.60$ & $5.00$ & $35.80$ & $8.66$ \\ \hline
w/ $\mathcal{R}_{\rm on}$  & $10^{-3}$  & $88.29$ & $3.52$ & $7.39$ & $6.83$
& $85.69$ & $4.43$ & $38.00$ & $9.01$ \\ \hline
w/ $\mathcal{R}_{\rm on}$  & $10^{-4}$  & $87.93$ & $4.48$ & $5.33$ & $4.83$  
& $85.12$ & $6.76$ & $43.87$ &  $5.95$\\ \hline
w/ $\mathcal{R}_{\rm on}$  & $10^{-5}$ & $87.61$ & $4.69$ & $3.83$ & $4.73$ 
& $85.39$ & $6.35$ & $35.70$ & $5.30$ \\ \hline
w/ Both
&  $10^{-4}$ & $87.44$ & $3.69$ & $9.69$ & $9.10$ & $84.53$ & $4.43$ & $63.92$ & $10.76$ \\ \bottomrule
\end{tabular}
\caption{Ablation study on the 20NG\textsubscript{15} and 20NG datasets. For OOD detection and misclassification detection, we report BAUCC$_{0.5}$. We set $\delta_y=0.1$ and $\delta_{\rm off}=10^{-3}$. }
\label{table:ablation}
\end{table*}

\section{Related Works and Discussion}

\noindent\textbf{Other Related Works:} \citet{Lakshminarayanan2017ensemble} propose a model ensembling approach to improve model calibration. They first train multiple models with different initializations and then average their predictions. However, fine-tuning multiple language models requires extremely intensive computing resources.

\citet{kumar2018trainable} propose a differentiable surrogate for the expected calibration error, called maximum mean calibration error (MMCE), using kernel embedding. However, such a kernel embedding method is computationally expensive and not scalable to the large  pre-trained language models.

\noindent\textbf{Accelerating Optimization:} To further improve the calibration performance of our method, we can leverage some recent minimax optimization techniques to better solve the two inner optimization problems in \eqref{eq:in-domain:x} and \eqref{eq:out-of-domain} without increasing the computational complexity. For example, \citet{zhang2019you} propose an efficient approximation algorithm based on Pontryagin's Maximal Principle to replace the multi-step projected gradient update for the inner optimization problem. Another option is the learning-to-learn framework \citep{DBLP:journals/corr/abs-1811-01213}, where the inner problem is solved by a learnt optimizer. These techniques can help us obtain $\mathbf{x}^{\prime}$ and $\mathbf{x}^{\prime\prime}$ more efficiently.

\noindent\textbf{Connection to Robustness:} The interpolated training samples
can naturally promote the local Lipschitz continuity of our model. Such a local
smoothness property has several advantages: (1) It makes the model more robust
to the inherent noise in the data, \eg, noisy labels; (2) it is particularly
helpful to prevent overfitting and improve generalization, especially for low-resource tasks.

\noindent\textbf{Extensions:} Our method is quite general and can be applied to other deep neural network-based problems besides language model fine-tuning.

\section{Conclusion}
We have proposed a regularization method to mitigate miscalibration of fine-tuned language models from a data augmentation perspective.
Our method imposes two new regularizers using generated on- and off- manifold samples to improve both in-distribution and out-of-distribution calibration.
Extensive experiments on six datasets demonstrate that  our method outperforms state-of-the-art calibration methods in terms of expected calibration error, misclassification detection and OOD detection.

\section*{Acknowledgement}

This work was supported in part by the National Science Foundation award III-2008334, Amazon Faculty Award, and Google Faculty Award.

\bibliographystyle{ims}
\bibliography{main}

\begin{thebibliography}{39}
\expandafter\ifx\csname natexlab\endcsname\relax\def\natexlab#1{#1}\fi
\expandafter\ifx\csname url\endcsname\relax
  \def\url#1{\texttt{#1}}\fi
\expandafter\ifx\csname urlprefix\endcsname\relax\def\urlprefix{}\fi

\bibitem[{Blundell et~al.(2015)Blundell, Cornebise, Kavukcuoglu and
  Wierstra}]{pmlr-v37-blundell15}
\textsc{Blundell, C.}, \textsc{Cornebise, J.}, \textsc{Kavukcuoglu, K.} and
  \textsc{Wierstra, D.} (2015).
\newblock Weight uncertainty in neural network.
\newblock In \textit{International Conference on Machine Learning}.

\bibitem[{Chang et~al.(2008)Chang, Ratinov, Roth and
  Srikumar}]{chang2008importance}
\textsc{Chang, M.-W.}, \textsc{Ratinov, L.}, \textsc{Roth, D.} and
  \textsc{Srikumar, V.} (2008).
\newblock Importance of semantic representation: Dataless classification.
\newblock In \textit{Proceedings of the Twenty-Third AAAI Conference on
  Artificial Intelligence}.

\bibitem[{Chouldechova(2017)}]{chouldechova2017fair}
\textsc{Chouldechova, A.} (2017).
\newblock Fair prediction with disparate impact: A study of bias in recidivism
  prediction instruments.
\newblock \textit{Big data}, \textbf{5} 153--163.

\bibitem[{Devlin et~al.(2019)Devlin, Chang, Lee and Toutanova}]{devlin2019bert}
\textsc{Devlin, J.}, \textsc{Chang, M.-W.}, \textsc{Lee, K.} and
  \textsc{Toutanova, K.} (2019).
\newblock Bert: Pre-training of deep bidirectional transformers for language
  understanding.
\newblock In \textit{Proceedings of the 2019 Conference of the North American
  Chapter of the Association for Computational Linguistics: Human Language
  Technologies, Volume 1 (Long and Short Papers)}.

\bibitem[{Gal and Ghahramani(2016)}]{gal2016dropout}
\textsc{Gal, Y.} and \textsc{Ghahramani, Z.} (2016).
\newblock Dropout as a bayesian approximation: Representing model uncertainty
  in deep learning.
\newblock In \textit{International Conference on Machine Learning}.

\bibitem[{Gal et~al.(2017)Gal, Islam and Ghahramani}]{gal2017active}
\textsc{Gal, Y.}, \textsc{Islam, R.} and \textsc{Ghahramani, Z.} (2017).
\newblock Deep bayesian active learning with image data.
\newblock In \textit{International Conference on Machine Learning}.

\bibitem[{Gilmer et~al.(2018)Gilmer, Metz, Faghri, Schoenholz, Raghu,
  Wattenberg, Goodfellow and Brain}]{gilmer2018relationship}
\textsc{Gilmer, J.}, \textsc{Metz, L.}, \textsc{Faghri, F.},
  \textsc{Schoenholz, S.~S.}, \textsc{Raghu, M.}, \textsc{Wattenberg, M.},
  \textsc{Goodfellow, I.} and \textsc{Brain, G.} (2018).
\newblock The relationship between high-dimensional geometry and adversarial
  examples.
\newblock \textit{arXiv preprint arXiv:1801.02774}.

\bibitem[{Guo et~al.(2017)Guo, Pleiss, Sun and Weinberger}]{guo2017calibration}
\textsc{Guo, C.}, \textsc{Pleiss, G.}, \textsc{Sun, Y.} and \textsc{Weinberger,
  K.~Q.} (2017).
\newblock On calibration of modern neural networks.
\newblock In \textit{International Conference on Machine Learning}.

\bibitem[{Hendrycks and Gimpel(2016)}]{hendrycks2016}
\textsc{Hendrycks, D.} and \textsc{Gimpel, K.} (2016).
\newblock A baseline for detecting misclassified and out-of-distribution
  examples in neural networks.
\newblock In \textit{International Conference on Learning Representations}.

\bibitem[{Jiang et~al.(2018)Jiang, Chen, Shi, Dai and
  Zhao}]{DBLP:journals/corr/abs-1811-01213}
\textsc{Jiang, H.}, \textsc{Chen, Z.}, \textsc{Shi, Y.}, \textsc{Dai, B.} and
  \textsc{Zhao, T.} (2018).
\newblock Learning to defense by learning to attack.
\newblock \textit{arXiv preprint arXiv:1811.01213}.

\bibitem[{Jiang et~al.(2020)Jiang, He, Chen, Liu, Gao and
  Zhao}]{jiang2019smart}
\textsc{Jiang, H.}, \textsc{He, P.}, \textsc{Chen, W.}, \textsc{Liu, X.},
  \textsc{Gao, J.} and \textsc{Zhao, T.} (2020).
\newblock {SMART}: Robust and efficient fine-tuning for pre-trained natural
  language models through principled regularized optimization.
\newblock In \textit{Proceedings of the 58th Annual Meeting of the Association
  for Computational Linguistics}.

\bibitem[{Kim(2014)}]{kim2014convolutional}
\textsc{Kim, Y.} (2014).
\newblock Convolutional neural networks for sentence classification.
\newblock In \textit{Proceedings of the 2014 Conference on Empirical Methods in
  Natural Language Processing (EMNLP)}.

\bibitem[{Kingma and Ba(2014)}]{kingma2014adam}
\textsc{Kingma, D.~P.} and \textsc{Ba, J.} (2014).
\newblock Adam: A method for stochastic optimization.
\newblock \textit{arXiv preprint arXiv:1412.6980}.

\bibitem[{Kowsari et~al.(2017)Kowsari, Brown, Heidarysafa, Jafari~Meimandi, ,
  Gerber and Barnes}]{kowsari2017HDLTex}
\textsc{Kowsari, K.}, \textsc{Brown, D.~E.}, \textsc{Heidarysafa, M.},
  \textsc{Jafari~Meimandi, K.}, , \textsc{Gerber, M.~S.} and \textsc{Barnes,
  L.~E.} (2017).
\newblock Hdltex: Hierarchical deep learning for text classification.
\newblock In \textit{IEEE International Conference on Machine Learning and
  Applications (ICMLA)}.

\bibitem[{Kumar et~al.(2018)Kumar, Sarawagi and Jain}]{kumar2018trainable}
\textsc{Kumar, A.}, \textsc{Sarawagi, S.} and \textsc{Jain, U.} (2018).
\newblock Trainable calibration measures for neural networks from kernel mean
  embeddings.
\newblock In \textit{International Conference on Machine Learning}.

\bibitem[{Lakshminarayanan et~al.(2017)Lakshminarayanan, Pritzel and
  Blundell}]{Lakshminarayanan2017ensemble}
\textsc{Lakshminarayanan, B.}, \textsc{Pritzel, A.} and \textsc{Blundell, C.}
  (2017).
\newblock Simple and scalable predictive uncertainty estimation using deep
  ensembles.
\newblock In \textit{Advances in Neural Information Processing Systems}.

\bibitem[{Lan et~al.(2020)Lan, Chen, Goodman, Gimpel, Sharma and
  Soricut}]{Lan2020ALBERT}
\textsc{Lan, Z.}, \textsc{Chen, M.}, \textsc{Goodman, S.}, \textsc{Gimpel, K.},
  \textsc{Sharma, P.} and \textsc{Soricut, R.} (2020).
\newblock Albert: A lite bert for self-supervised learning of language
  representations.
\newblock In \textit{International Conference on Learning Representations}.
\newline\urlprefix\url{https://openreview.net/forum?id=H1eA7AEtvS}

\bibitem[{Lee et~al.(2018)Lee, Lee, Lee and Shin}]{lee2018simple}
\textsc{Lee, K.}, \textsc{Lee, K.}, \textsc{Lee, H.} and \textsc{Shin, J.}
  (2018).
\newblock A simple unified framework for detecting out-of-distribution samples
  and adversarial attacks.
\newblock In \textit{Advances in Neural Information Processing Systems}.

\bibitem[{Liu et~al.(2019)Liu, Ott, Goyal, Du, Joshi, Chen, Levy, Lewis,
  Zettlemoyer and Stoyanov}]{liu2019roberta}
\textsc{Liu, Y.}, \textsc{Ott, M.}, \textsc{Goyal, N.}, \textsc{Du, J.},
  \textsc{Joshi, M.}, \textsc{Chen, D.}, \textsc{Levy, O.}, \textsc{Lewis, M.},
  \textsc{Zettlemoyer, L.} and \textsc{Stoyanov, V.} (2019).
\newblock {RoBERTa}: {A} robustly optimized {BERT} pretraining approach.
\newblock \textit{arXiv preprint arXiv:1907.11692}.

\bibitem[{Louizos and Welling(2017)}]{pmlr-v70-louizos17a}
\textsc{Louizos, C.} and \textsc{Welling, M.} (2017).
\newblock Multiplicative normalizing flows for variational {B}ayesian neural
  networks.
\newblock In \textit{International Conference on Machine Learning}.

\bibitem[{Miyato et~al.(2018)Miyato, Maeda, Koyama and
  Ishii}]{miyato2018virtual}
\textsc{Miyato, T.}, \textsc{Maeda, S.-i.}, \textsc{Koyama, M.} and
  \textsc{Ishii, S.} (2018).
\newblock Virtual adversarial training: a regularization method for supervised
  and semi-supervised learning.
\newblock \textit{IEEE transactions on pattern analysis and machine
  intelligence}, \textbf{41} 1979--1993.

\bibitem[{M{\"u}ller et~al.(2019)M{\"u}ller, Kornblith and
  Hinton}]{muller2019does}
\textsc{M{\"u}ller, R.}, \textsc{Kornblith, S.} and \textsc{Hinton, G.~E.}
  (2019).
\newblock When does label smoothing help?
\newblock In \textit{Advances in Neural Information Processing Systems}.

\bibitem[{Naeini et~al.(2015)Naeini, Cooper and
  Hauskrecht}]{naeini2015obtaining}
\textsc{Naeini, M.~P.}, \textsc{Cooper, G.~F.} and \textsc{Hauskrecht, M.}
  (2015).
\newblock Obtaining well calibrated probabilities using bayesian binning.
\newblock In \textit{Proceedings of the Twenty-Ninth AAAI Conference on
  Artificial Intelligence}.

\bibitem[{Niculescu-Mizil and Caruana(2005)}]{mizil-2005-predict}
\textsc{Niculescu-Mizil, A.} and \textsc{Caruana, R.} (2005).
\newblock Predicting good probabilities with supervised learning.
\newblock In \textit{International Conference on Machine Learning}.

\bibitem[{Pereyra et~al.(2017)Pereyra, Tucker, Chorowski, Kaiser and
  Hinton}]{Pereyra2017erl}
\textsc{Pereyra, G.}, \textsc{Tucker, G.}, \textsc{Chorowski, J.},
  \textsc{Kaiser, {\L}.} and \textsc{Hinton, G.} (2017).
\newblock Regularizing neural networks by penalizing confident output
  distributions.
\newblock \textit{arXiv preprint arXiv:1701.06548}.

\bibitem[{Raffel et~al.(2019)Raffel, Shazeer, Roberts, Lee, Narang, Matena,
  Zhou, Li and Liu}]{raffel2019exploring}
\textsc{Raffel, C.}, \textsc{Shazeer, N.}, \textsc{Roberts, A.}, \textsc{Lee,
  K.}, \textsc{Narang, S.}, \textsc{Matena, M.}, \textsc{Zhou, Y.}, \textsc{Li,
  W.} and \textsc{Liu, P.~J.} (2019).
\newblock Exploring the limits of transfer learning with a unified text-to-text
  transformer.
\newblock \textit{arXiv preprint arXiv:1910.10683}.

\bibitem[{Shen et~al.(2018)Shen, Yun, Lipton, Kronrod and
  Anandkumar}]{shen2018deep}
\textsc{Shen, Y.}, \textsc{Yun, H.}, \textsc{Lipton, Z.~C.}, \textsc{Kronrod,
  Y.} and \textsc{Anandkumar, A.} (2018).
\newblock Deep active learning for named entity recognition.
\newblock In \textit{International Conference on Learning Representations}.

\bibitem[{Siddhant and Lipton(2018)}]{siddhant2018deep}
\textsc{Siddhant, A.} and \textsc{Lipton, Z.~C.} (2018).
\newblock Deep bayesian active learning for natural language processing:
  Results of a large-scale empirical study.
\newblock In \textit{Proceedings of the 2018 Conference on Empirical Methods in
  Natural Language Processing}.

\bibitem[{Socher et~al.(2012)Socher, Bengio and Manning}]{socher2012deep}
\textsc{Socher, R.}, \textsc{Bengio, Y.} and \textsc{Manning, C.~D.} (2012).
\newblock Deep learning for nlp (without magic).
\newblock In \textit{Tutorial Abstracts of ACL 2012}.

\bibitem[{Stutz et~al.(2019)Stutz, Hein and Schiele}]{stutz2019disentangling}
\textsc{Stutz, D.}, \textsc{Hein, M.} and \textsc{Schiele, B.} (2019).
\newblock Disentangling adversarial robustness and generalization.
\newblock In \textit{Proceedings of the IEEE Conference on Computer Vision and
  Pattern Recognition}.

\bibitem[{Thulasidasan et~al.(2019)Thulasidasan, Chennupati, Bilmes,
  Bhattacharya and Michalak}]{sunil2019mixup}
\textsc{Thulasidasan, S.}, \textsc{Chennupati, G.}, \textsc{Bilmes, J.~A.},
  \textsc{Bhattacharya, T.} and \textsc{Michalak, S.} (2019).
\newblock On mixup training: Improved calibration and predictive uncertainty
  for deep neural networks.
\newblock In \textit{Advances in Neural Information Processing Systems}.

\bibitem[{Verma et~al.(2019)Verma, Lamb, Beckham, Najafi, Mitliagkas, Lopez-Paz
  and Bengio}]{verma-2019-manifold}
\textsc{Verma, V.}, \textsc{Lamb, A.}, \textsc{Beckham, C.}, \textsc{Najafi,
  A.}, \textsc{Mitliagkas, I.}, \textsc{Lopez-Paz, D.} and \textsc{Bengio, Y.}
  (2019).
\newblock Manifold mixup: Better representations by interpolating hidden
  states.
\newblock In \textit{International Conference on Machine Learning}.

\bibitem[{Wang et~al.(2019)Wang, Pruksachatkun, Nangia, Singh, Michael, Hill,
  Levy and Bowman}]{wang2019superglue}
\textsc{Wang, A.}, \textsc{Pruksachatkun, Y.}, \textsc{Nangia, N.},
  \textsc{Singh, A.}, \textsc{Michael, J.}, \textsc{Hill, F.}, \textsc{Levy,
  O.} and \textsc{Bowman, S.} (2019).
\newblock Superglue: A stickier benchmark for general-purpose language
  understanding systems.
\newblock In \textit{Advances in Neural Information Processing Systems}.

\bibitem[{Wang et~al.(2018)Wang, Singh, Michael, Hill, Levy and
  Bowman}]{wang2018glue}
\textsc{Wang, A.}, \textsc{Singh, A.}, \textsc{Michael, J.}, \textsc{Hill, F.},
  \textsc{Levy, O.} and \textsc{Bowman, S.~R.} (2018).
\newblock Glue: A multi-task benchmark and analysis platform for natural
  language understanding.
\newblock In \textit{International Conference on Learning Representations}.

\bibitem[{Wilcoxon(1992)}]{wilcoxon1992individual}
\textsc{Wilcoxon, F.} (1992).
\newblock Individual comparisons by ranking methods.
\newblock In \textit{Breakthroughs in statistics}. Springer, 196--202.

\bibitem[{Wong et~al.(2019)Wong, Rice and Kolter}]{wong2020fast}
\textsc{Wong, E.}, \textsc{Rice, L.} and \textsc{Kolter, J.~Z.} (2019).
\newblock Fast is better than free: Revisiting adversarial training.
\newblock In \textit{International Conference on Learning Representations}.

\bibitem[{Zhang et~al.(2019)Zhang, Zhang, Lu, Zhu and Dong}]{zhang2019you}
\textsc{Zhang, D.}, \textsc{Zhang, T.}, \textsc{Lu, Y.}, \textsc{Zhu, Z.} and
  \textsc{Dong, B.} (2019).
\newblock You only propagate once: Accelerating adversarial training via
  maximal principle.
\newblock In \textit{Advances in Neural Information Processing Systems}.

\bibitem[{Zhang et~al.(2018)Zhang, Cisse, Dauphin and
  Lopez-Paz}]{zhang2018mixup}
\textsc{Zhang, H.}, \textsc{Cisse, M.}, \textsc{Dauphin, Y.~N.} and
  \textsc{Lopez-Paz, D.} (2018).
\newblock mixup: Beyond empirical risk minimization.
\newblock In \textit{International Conference on Learning Representations}.

\bibitem[{Zhang et~al.(2015)Zhang, Zhao and LeCun}]{zhang2015character}
\textsc{Zhang, X.}, \textsc{Zhao, J.} and \textsc{LeCun, Y.} (2015).
\newblock Character-level convolutional networks for text classification.
\newblock In \textit{Advances in neural information processing systems}.

\end{thebibliography}
\newpage
\appendix
\section{Dataset Details}
\label{sec:app:dataset}
\begin{table}[!htb]
\centering
\begin{tabular}{@{}lcccc@{}}
\toprule
       & \multicolumn{1}{l}{\#Train} & \multicolumn{1}{l}{\#Dev} & \multicolumn{1}{l}{\#Test} & \multicolumn{1}{l}{\#Label} \\ \midrule
20NG$_{15}$ & 7010                        & 1753                      & 5833                       & 15                          \\
20NG$_5$  & -                           & -                         & 1699                       & 5                           \\ \hline
20NG   & 9051                        & 2263                      & 7532                       & 20                          \\
SST-2  & -                           & -                         & 1822                       & 2                           \\\hline
WOS$_{100}$ & 16794                       & 4191                      & 13970                      & 100                         \\
WOS$_{34}$  & -                           & -                         & 4824                       & 34                          \\\hline
WOS    & 22552                       & 5639                      & 18794                      & 134                         \\
AGnews & -                           & -                         & 7600                       & 4                           \\\hline
Yahoo$_8$ & 16000                       & 4000                      & 48000                      & 8                           \\
Yahoo$_2$ & -                           & -                         & 12000                      & 2                           \\\hline
Yahoo  & 20000                       & 5000                      & 60000                      & 10                          \\
Yelp   & -                           & -                         & 38000                      & 2                           \\ \bottomrule
\end{tabular}
\caption{Dataset statistics and dataset split. '-' denotes that this part is not used. The original Yahoo dataset contains $140,000$ training samples for each class which is too large; we randomly draw $2,000$ and $500$ samples for each class as our training and development set. }
\label{table:dataset}
\end{table}

All the data are publicly available. We also offer the links to the data as follows:
\begin{enumerate}
    \item 20NG: \url{http://qwone.com/~jason/20Newsgroups/}.
    \item SST-2: \url{https://nlp.stanford.edu/sentiment/index.html}.
    \item WOS: \url{https://data.mendeley.com/datasets/9rw3vkcfy4/2}.
    \item AGnews: \url{https://github.com/yumeng5/WeSTClass}.
    \item Yahoo: \url{https://www.kaggle.com/soumikrakshit/yahoo-answers-dataset}.
    \item Yelp: \url{https://github.com/yumeng5/WeSTClass}.
\end{enumerate}
\section{Experiment Details}
\label{app:exp_details}
We use ADAM \citep{kingma2014adam} with $\beta_1=0.9$ and $\beta_2=0.999$ as the optimizer in all the datasets. We use the learning rate of $5\times 10^{-5}$ and batch size $32$ except $1\times10^{-5}$ and $16$ for Yahoo\textsubscript{8} and Yahoo. We set the maximum number of epochs to 5 in Yahoo\textsubscript{8} and Yahoo and $10$ in the other datasets. We use the dropout rate of $0.1$ as in \citep{devlin2019bert}. The documents are tokenized using wordpieces and are chopped to spans no longer than $150$ tokens on 20NG$_{15}$ and 20NG and 256 on other datasets..

\noindent\textbf{Hyper-parameters:} For our method, we use $\lambda_{\rm on}=\lambda_{\rm off}=1$, $\delta_{\rm on}=10^{-4}$, $\delta_{\rm off}=10^{-3}$ and $\delta_y =0.1$ for all the datasets. We then conduct an extensive hyper-parameter search for the baselines: for label smoothing, we search the smoothing parameter from $\{0.05, 0.1 \}$ as in \citep{muller2019does}; for ERL, the penalty weight is chosen from $\{0.05, 0.1, 0.25, 0.5, 1, 2.5, 5\}$; for VAT, we search the perturbation size in $\{10^{-3}, 10^{-4}, 10^{-5}\}$ as in \citep{jiang2019smart}; for Mixup, we search the interpolation parameter from $\{0.1,0.2,0.3,0.4 \}$ as suggested in \citep{zhang2018mixup,sunil2019mixup}; for Manifold-mixup, we search from $\{0.2,0.4,1,2,4 \}$.  We perform 10 stochastic forward passes for MCDP at test time. For hyper-parameter tuning, we run all the methods 5 times and then take the average. The hyper-parameters are selected to get the best ECE on the development set of each dataset. The interpolation of Mixup is performed on the input embeddings obtained from the first layer of the language model; the interpolation of Manifold-mixup is performed on the features obtained from the last layer of the language model.

\label{sec:appendix}

\section{Metrics of Misclassification and Out-of-distribution detection}
\label{sec:appendix:metric}

Existing works on out-of-distribution (OOD) detection and misclassification detection \citep{hendrycks2016} use 
traditional binary classification metrics, \eg, AUPR and AUROC.
As we discussed in Section 1 and 2, the output probability of a calibrated model should reflect the true likelihood.
However, AUROC and AUPR cannot reflect true model calibration because the model can still achieve high scores even though it has high confidences for misclassified and OOD samples.
We argue that it is more reasonable to use the Normalized Bounded Area Under the Calibration Curve (NBAUCC) defined as in Section 4.

\begin{table*}[h]
\small
    \centering
    \begin{tabular}{c|cccc|c|c|c|c|c}
        \multirow{2}{*}{Model}& \multicolumn{4}{c|}{Confidence} & \multirow{2}{*}{Optimal $\tau$} & \multirow{2}{*}{AUPR} & \multirow{2}{*}{AUROC} & \multirow{2}{*}{${\rm NBAUCC}_{1}$}&\multirow{2}{*}{${\rm NBAUCC}_{0.5}$}\\
        \cline{2-5}
         & $x_{\rm in, 1}$ & $x_{\rm in, 2}$ & $x_{\rm out, 1}$ & $x_{\rm out, 2}$ &&&&\\
        \hline
        $h_1$ (Miscalibrated) & 0.9 & 0.95 & 0.8 & 0.85 & $(0.85, 0.9)$ & 0.417 & 1 & 0.145&0 \\
        $h_2$ (Well-calibraterd) & 0.9 & 0.95 & 0.1 & 0.15 & $(0.15, 0.9)$ & 0.417 & 1 & 0.845&0.773
    \end{tabular}
    \caption{NBAUCC vs. AUROC/AUPR}
    \label{tab:aucc_example}
\end{table*}

Table~\ref{tab:aucc_example} shows an illustrative example. 
As can be seen, $h_1$ is better calibrated than $h_2$, since $h_1$ can detect OOD samples under a wide range of threshold ($0.15 < \tau < 0.9$) while $h_2$ requires an absurdly large threshold ($0.85 < \tau < 0.9$). 
However, if we use the traditional AUPR and AUROC metrics, we will conclude that $h_1$ is as well calibrated as $h_2$ since AUPR\textsuperscript{$h_1$} = AUPR\textsuperscript{$h_2$} = $0.417$ and  AUROC\textsuperscript{$h_1$} = AUROC\textsuperscript{$h_2$}= $1$. 
On the other hand, if we use NBAUCC, we will have ${\rm NBAUCC}^{h_1}_{1} =0.845 >  {\rm NBAUCC}^{h_1}_{1} = 0.145$, or ${\rm NBAUCC}^{h_1}_{0.5} =0.773 >  {\rm NBAUCC}^{h_1}_{0.5} = 0$  which can reflect the true calibration of the two classifiers.

We remark that it is more appropriate to use ${\rm NBAUCC}_{0.5}$ than ${\rm NBAUCC}_{1}$ since a calibrated model should provide low confidences for the misclassified and OOD samples and it is unreasonable to use a large threshold to detect them.

\section{Additional Results}
\label{sec:appendix:tau}
\begin{figure*}[h]
 \centering
 \includegraphics[width=\linewidth]{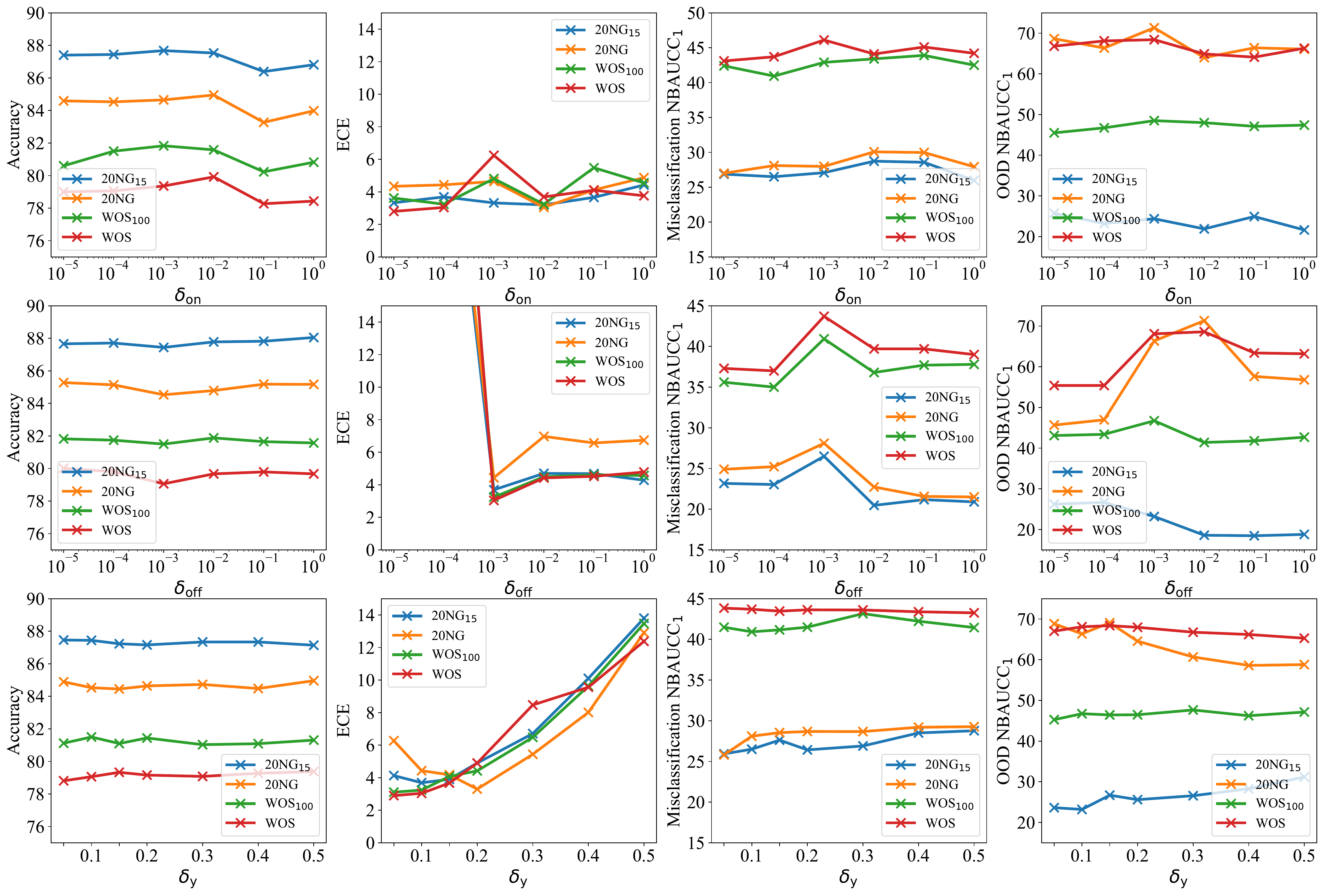}
 \caption{ Parameter study of $\delta_{\rm on}$, $\delta_{\rm off}$ and $\delta_y$. We use ${\rm NBAUCC}_{1}$ for OOD and misclassification detection.}
 \label{fig:para2}
\end{figure*}

\begin{figure*}[h]
 \centering
 \includegraphics[width=\linewidth]{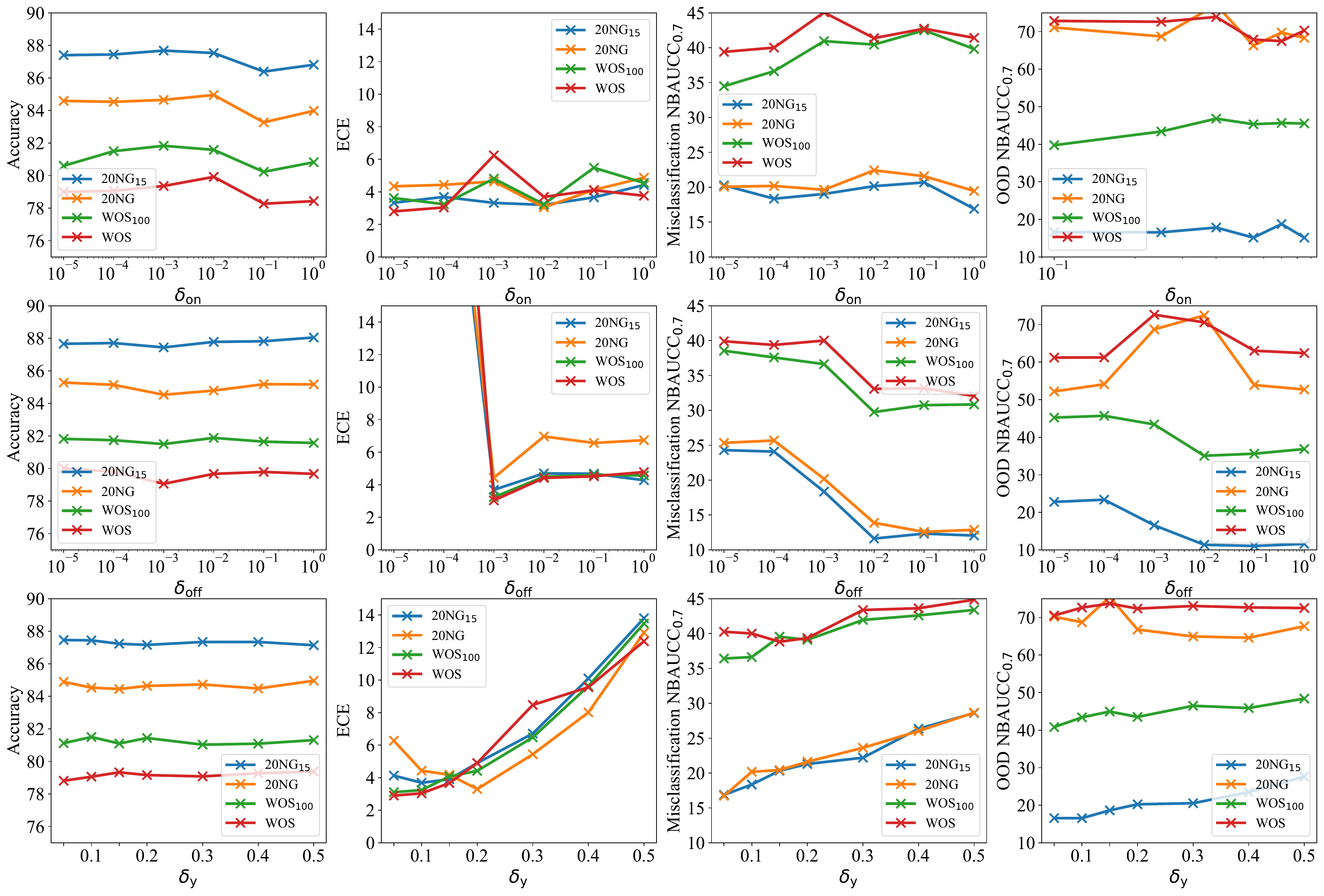}
 \caption{ Parameter study of $\delta_{\rm on}$, $\delta_{\rm off}$ and $\delta_y$. We use ${\rm NBAUCC}_{0.7}$ for OOD and misclassification detection. }
 \label{fig:para3}
\end{figure*}

Table~\ref{table:mis_ood_new} and \ref{table:mis_ood_new2} report the NBAUCCs of all the methods on misclassification and OOD detection when $
\tau_{\rm upper}=0.7$ and $
\tau_{\rm upper}=1$. Table~\ref{table:ablation_new} and \ref{table:ablation_new2} report the ablation study results of all the methods when $
\tau_{\rm upper}=0.7$ and $
\tau_{\rm upper}=1$. Figure~\ref{fig:para2} and \ref{fig:para3} report the parameter study results of all the methods when $
\tau_{\rm upper}=0.7$ and $
\tau_{\rm upper}=1$.

\begin{table*}[!h]
\centering
\begin{tabular}{@{}l@{}c@{~}c@{~}c@{~}c@{~}c@{~}c@{~}|@{~}c@{~}@{~}c@{~}c@{~}c@{~}c@{~}c@{~}c@{~}c@{}}
\toprule
&& \multicolumn{4}{c}{Misclassification Detection} && \multicolumn{6}{c}{OOD Detection}
\\
\cline{2-13}

Data & \multirow{2}{*}{20NG\textsubscript{15}} & \multirow{2}{*}{20NG} & \multirow{2}{*}{WOS\textsubscript{100}} & \multirow{2}{*}{WOS} & \multirow{2}{*}{Yahoo\textsubscript{8}} & \multirow{2}{*}{Yahoo}
& 20NG\textsubscript{15} & 20NG & WOS\textsubscript{100} & WOS & Yahoo\textsubscript{8} & Yahoo \\

( OOD )&  &  &  &  &  &
& 20NG\textsubscript{5} & SST-2 & WOS\textsubscript{34} & AGnews & Yahoo\textsubscript{2} & Yelp \\ \hline
BERT
& 17.86 & 18.48 & 35.84 & 39.08 & 28.83 & 29.67
& 13.52 & 42.86 & 40.04 & 59.42 & 26.63 & 38.30 \\
TS
& 23.74 & 23.58 & 38.34 & 40.76 & 31.10 & 32.63
& 19.74 & 50.00 & 42.96 & 60.70 & 28.30 & 42.07\\
MCDP
& 23.58 & 24.58 & 38.54 & 41.20 & 31.43 & 32.57
& 16.82 & 44.96 & 42.74 & 60.72 & 27.47 & 39.83\\
LS
& 21.22 & 23.24 & 37.22 & 40.12 & 30.93 & 34.30
& 18.76 & 55.24 & 42.54 & 63.62 & 27.87 & 40.77\\
ERL
& 24.04 & 25.68 & 37.87 & 41.17 & 32.27 & 33.90
& 22.10 & 54.20 & 42.67 & 62.10 & 28.73 & 43.37\\
VAT
& 17.80 & 17.50 & 35.90 & 38.80 & 27.87 & 31.13
& 13.00 & 49.00 & 40.30 & 62.50 & 25.80 & 40.63\\
Mixup
& 21.42 & 21.86 & 37.72 & 40.92 & 30.97 & 32.97
& 16.70 & 50.94 & 42.13 & 62.98 & 28.00 & 44.57\\
M-mixup
& 17.86 & 19.24 & 36.48 & 38.33 & 29.67 & 31.50
& 14.06 & 44.56 & 41.51 & 61.30 & 27.43 & 44.20\\
\hline
\textbf{Ours}
& \textbf{26.50} & \textbf{28.10} & \textbf{40.93} & \textbf{43.70} & \textbf{33.07} & \textbf{35.13}
& \textbf{23.20} & \textbf{66.36} & \textbf{46.73} & \textbf{68.10} & \textbf{29.70} & \textbf{46.43}
\\ \bottomrule
\end{tabular}
\caption{$\rm NBAUCC_{1}$ on misclassification detection and OOD detection. We report the average performance of 5 random initializations.}
\label{table:mis_ood_new}
\end{table*}

\begin{table*}[h]
\centering
\begin{tabular}{@{}l@{}c@{~}c@{~}c@{~}c@{~}c@{~}c@{~}|@{~}c@{~}@{~}c@{~}c@{~}c@{~}c@{~}c@{~}c@{~}c@{}}
\toprule
&& \multicolumn{4}{c}{Misclassification Detection} && \multicolumn{6}{c}{OOD Detection}
\\
\cline{2-13}

Data & \multirow{2}{*}{20NG\textsubscript{15}} & \multirow{2}{*}{20NG} & \multirow{2}{*}{WOS\textsubscript{100}} & \multirow{2}{*}{WOS} & \multirow{2}{*}{Yahoo\textsubscript{8}} & \multirow{2}{*}{Yahoo} 
& 20NG\textsubscript{15} & 20NG & WOS\textsubscript{100} & WOS & Yahoo\textsubscript{8} & Yahoo \\
 
( OOD )&  &  &  &  &  &  
& 20NG\textsubscript{5} & SST-2 & WOS\textsubscript{34} & AGnews & Yahoo\textsubscript{2} & Yelp \\ \hline
BERT
& 8.26 & 8.70  & 26.95 & 31.18 & 18.52 & 19.46 
& 7.05 & 33.24  & 32.97 & 57.45 & 18.86 & 27.68 \\ 
TS 
& 14.60 & 13.72  & 31.73 & 33.89 & 22.32 & 24.61 
& 12.91 & 43.55  & 37.84 & 59.86 & 22.17 & 34.03 \\ 
MCDP
& 13.14 & 14.21  & 31.05 & 34.74 & 21.41 & 22.62 
& 9.85 & 36.96  & 36.97 & 60.06 & 19.99 & 29.45 \\ 
LS 
& 12.45 & 14.24  & 30.92 & 33.51 & 22.94 & 27.52
& 11.63 & 49.60  & 36.04 & 65.28 & 22.38 & 33.00 \\ 
ERL
& 17.92 & 20.04  & 30.83 & 35.26 & 25.07 & 27.34 
& 15.43 & 55.69  & 36.69 & 61.93 & 24.07 & 36.74 \\ 
VAT 
& 8.44 & 9.66  & 29.39 & 30.57 & 17.23 & 21.74 
& 7.26 & 41.35  & 32.56 & 60.81 & 17.64 & 31.17 \\  
Mixup 
& 13.33 & 11.87  & 31.71 & 35.24 & 22.62 & 22.80 
& 11.50 & 43.60  & 37.09 & 65.51 & 22.19 & 33.66 \\  
M-mixup 
& 8.67 & 9.89  & 27.33 & 29.61 & 20.33 & 23.05 
& 7.18 & 37.10  & 33.57 & 58.13 & 20.66 & 36.42\\ 
\hline
\textbf{Ours} 
& \textbf{18.35} & \textbf{20.18} & \textbf{36.63} & \textbf{40.01} & \textbf{25.94} & \textbf{29.15}
& \textbf{16.55} & \textbf{68.72} & \textbf{43.40} & \textbf{72.62} & \textbf{25.03} & \textbf{41.11} 
\\ \bottomrule
\end{tabular}
\caption{${\rm NBAUCC}_{0.7}$ on misclassification detection and OOD detection. We report the average performance of 5 random initializations.}
\label{table:mis_ood_new2}
\end{table*}

\begin{table*}[!htb]
\centering
\begin{tabular}{lc|cccc|cccc}
\toprule
 \multicolumn{2}{c|}{Dataset}& \multicolumn{4}{c|}{20NG\textsubscript{15}} & \multicolumn{4}{c}{20NG} \\ \hline
Model &$\delta_{\rm on}$ & Accuracy & ECE & OOD & Mis & Accuracy & ECE & OOD & Mis \\ \hline
BERT & -   & 87.42 & 9.24 & 13.52 & 17.86
& 84.55 & 11.61 & 42.86 & 18.48\\ \hline
w/ $\mathcal{R}_{\rm off}$ & - & 86.48 & 6.51 & 18.10 & 24.53 & 83.90 & 7.98 & 63.73 & 25.40 \\ \hline
w/ $\mathcal{R}_{\rm on}$  & $10^{-2}$   & 88.73 & 2.77 & 22.83 & 27.40 & 85.60 & 5.00 & 51.53 & 27.40 \\ \hline
w/ $\mathcal{R}_{\rm on}$  & $10^{-3}$  & 88.29 & 3.52 & 21.03 & 24.13 & 85.69 & 4.43 & 53.87 & 26.30 \\ \hline
w/ $\mathcal{R}_{\rm on}$  & $10^{-4}$  & 87.93 & 4.48 & 17.43 & 21.63 & 85.12 & 6.76 & 57.47 & 21.93 \\ \hline
w/ $\mathcal{R}_{\rm on}$  & $10^{-5}$ & 87.61 & 4.69 & 15.73 & 21.43 & 85.39 & 6.35 & 52.07 & 21.63 \\ \hline
w/ Both
&  $10^{-4}$ & 87.44 & 3.69 & 23.20 & 26.50 & 84.53 & 4.43 & 66.36 & 28.10 \\ \bottomrule
\end{tabular}
\caption{Ablation study on the 20NG\textsubscript{15} and 20NG datasets. For OOD detection and misclassification detection, we report $\rm NBAUCC_{1}$. We set $\delta_y=0.1$ and $\delta_{\rm off}=10^{-3}$. }
\label{table:ablation_new}
\end{table*}

\begin{table*}[!htb]
\centering
\begin{tabular}{lc|cccc|cccc}
\toprule
 \multicolumn{2}{c|}{Dataset}& \multicolumn{4}{c|}{20NG\textsubscript{15}} & \multicolumn{4}{c}{20NG} \\ \hline
Model &$\delta_{\rm on}$ & Accuracy & ECE & OOD & Mis & Accuracy & ECE & OOD & Mis \\ \hline
BERT & -   & 87.42 & 9.24 & 7.05 & 8.26
& 84.55 & 11.61 & 33.24 & 8.70\\ \hline
w/ $\mathcal{R}_{\rm off}$ & - & 86.48 & 6.51 & 11.75 & 14.79
& 83.90 & 7.98 & 62.67 & 15.42 \\ \hline
w/ $\mathcal{R}_{\rm on}$  & $10^{-2}$   & 88.73 & 2.77 & 15.27  & 18.35  
& 85.60 & 5.00 & 46.67 & 18.39 \\ \hline
w/ $\mathcal{R}_{\rm on}$  & $10^{-3}$  & 88.29 & 3.52 & 13.86 & 15.66
& 85.69 & 4.43 & 50.07 & 18.17 \\ \hline
w/ $\mathcal{R}_{\rm on}$  & $10^{-4}$  & 87.93 & 4.48 & 10.61 & 12.59  
& 85.12 & 6.76 & 53.64 &  13.18\\ \hline
w/ $\mathcal{R}_{\rm on}$  & $10^{-5}$ & 87.61 & 4.69 & 8.71 & 12.25 
& 85.39 & 6.35 & 46.24 & 12.20 \\ \hline
w/ Both
&  $10^{-4}$ & 87.44 & 3.69 & 16.55 & 18.35 & 84.53 & 4.43 & 68.72 & 20.18 \\ \bottomrule
\end{tabular}
\caption{Ablation study on the 20NG\textsubscript{15} and 20NG datasets. For OOD detection and misclassification detection, we report NBAUCC$_{0.7}$. We set $\delta_y=0.1$ and $\delta_{\rm off}=10^{-3}$. }
\label{table:ablation_new2}
\end{table*}

\end{document}